\RequirePackage{fix-cm}
\documentclass[smallextended]{svjour3}       
\smartqed  
\usepackage{graphicx}
\usepackage{booktabs}
\usepackage[numbers,sort]{natbib}
\usepackage{marvosym}

\journalname{Preprint to arXiv} 
\begin{document}

\title{A new approach for generation of generalized basic probability assignment in the evidence theory
}


\author{Dongdong Wu         \and
        Zijing Liu          \and
        Yongchuan Tang        
}


\institute{Dongdong Wu  \at
           School of Big Data and Software Engineering, Chongqing University, Chongqing 401331, China\\
              \email{20171755@cqu.edu.cn}
           \and
           Zijing Liu  \at
           School of Big Data and Software Engineering, Chongqing University, Chongqing 401331, China\\
              \email{20171724@cqu.edu.cn}
           \and
           Yongchuan Tang (\Letter) \at
           School of Big Data and Software Engineering, Chongqing University, Chongqing 401331, China\\
              \email{tangyongchuan@cqu.edu.cn}
}

\date{Received: date / Accepted: date}

\maketitle

\begin{abstract}
The process of information fusion needs to deal with a large number of uncertain information with multi-source, heterogeneity, inaccuracy, unreliability, and incompleteness.
In practical engineering applications, Dempster-Shafer evidence theory is widely used in multi-source information fusion owing to its effectiveness in data fusion.
Information sources have an important impact on multi-source information fusion in an environment of complex, unstable, uncertain, and incomplete characteristics.
To address multi-source information fusion problem, this paper considers the situation of uncertain information modeling from the closed world to the open world assumption and studies the generation of basic probability assignment (BPA) with incomplete information.
In this paper, a new method is proposed to generate generalized basic probability assignment (GBPA) based on the triangular fuzzy number model under the open world assumption.
The proposed method can not only be used in different complex environments simply and flexibly, but also have less information loss in information processing.
Finally, a series of comprehensive experiments basing on the UCI data sets are used to verify the rationality and superiority of the proposed method.

\end{abstract}

\section{Introduction}
\label{Introduction}
With the development of artificial intelligence, the multi-source information fusion technology that can fuse the collected information to improve the security and stability of the system, is widely used in the real world \cite{zhang2017perceiving,zhou2017online}, such as decision-making \cite{debnath2018game,he2018evidential}, fault diagnosis \cite{janghorbani2017fuzzy,lin2018multisensor}, risk and reliability analysis \cite{song2018sensor,chemweno2018risk,dutta2017modeling,seiti2018developing,chen2018new}, 
pattern recognition \cite{liu2017combination,liu2017classifier}, etc.
However, in complex environments, the data received by sensors based on physical hardware technology is often inaccurate and uncertain. If the data are directly fused, the results may be incorrect or counter-intuitive.
The reason for such a result is often the interference information generated by the complex environment or the malfunctioning sensor recognition function. Therefore, how to effectively solve the fusion of uncertain information in complex environments is still an open issue.
To address this issue, many mathematical theories are proposed by scholars, like fuzzy set theory \cite{chen2013fuzzy}, belief function theory \cite{frikha2015analytic}, grey theory \cite{chen2018new}, probability theory \cite{nakamori2003linear,mcclean2001aggregation}, etc. There are also some methods basing on  Z-numbers \cite{kang2018stable,zadeh2011note}, D-numbers \cite{li2018d,mo2018new,xiao2018novel,bian2018failure} and so on \cite{XU201743}.

As a generalization of Bayesian theory \cite{feller2008introduction,bernardo2009bayesian}, Dempster-Shafer evidence theory was first proposed by Dempster in 1967 \cite{dempster1967upper} and developed by Shafer in 1976 \cite{shafer1976mathematical}.
As Dempster-Shafer evidence theory can flexibly handle information with incompleteness and uncertainty \cite{zhang2019weighted,8906062}, it is widely used in practical application systems \cite{han2018enhanced}, such as classification \cite{liu2017weighted}, clustering \cite{Zhou2017SELP,su2018bpec,meng2020belief}, reliability analysis \cite{deng2018dependence}, correlation analysis \cite{jiang2018correlation}, multi-attribute decision analysis \cite{fu2016distributed,fu2018determining}, fault diagnosis \cite{xiao2017novel,GONG2018395,zhang2019weighted,cheng2016new} and so on \cite{Su2019}.
Nevertheless, Dempster-Shafer evidence theory has many limitations that need to be improved. Firstly, the key  step is how to automatically generate basic probability assignment (BPA), which is the prerequisites for the application of Dempster-Shafer evidence theory and the use of Dempster's combination rule. However, the method for BPA generation is still a difficult and open issue.
The quality of generated BPA will directly affect subsequent practical applications. If the method of generating BPA is unreasonable, causing failure to generate an effective BPA, the fusion result is likely to be counter-intuitive.
Secondly, Yager et al. \cite{yager1987dempster} believed the normalization of the combination rule is the main reason for the unreasonable fusion results, so modification of Dempster's combination rule is suggested. Smets and Kennes \cite{smets1994transferable} pointed out that modifying the combination rule usually breaks good properties, like commutativity and associativity. And they proposed that it is unreasonable to modify the combination rule if the result is caused by a sensor failure. Therefore, the concepts of the open world and the closed world are introduced in the Transferable Belief Model (TBM) \cite{smets1994transferable}.
Dubois and Prade \cite{dubois1988default} suggest assigning the conflicting evidence to the union of focal elements.
Sun et al. \cite{sun2000new} proposed that even if there is a conflict between evidences, the evidences are also partly available. Therefore, by calculating the average of the conflict between every two pieces of evidences, they defined the valid coefficients of evidences, and then allocated total evidential conflict  to propositions according to a certain proportion. Li et al. \cite{li2001efficient} pointed out that Sun et al.'s method \cite{sun2000new} has a clear physical meaning and subjective factors of the formulas, but they are not obvious.
Some models are presented to allocate evidential conflict in accordance with the weighted averaging support of each proposition \cite{wang2019base,li2020improved}.
The belief entropy also attracts many attentions \cite{deng2016deng,song2018evidence,jirouvsek2018new,Zheng2020Deng}.
Other solutions for this problem can be found in \cite{xiao2018improved,xiao2019multi,liu2017weighted}.
In this paper, we will focus on the  process of using multi-source information to generate BPAs in the open world, which is the base of data fusion in the evidence theory.

Many BPA generation methods based on the complete frame of discernment have been proposed by researchers.
For example, Xu et al. \cite{xu2013new} generated BPA of a nested structure by probability density function. Denoeux \cite{denoeux2008k} used the training samples to generate BPA. The similarity between the training samples and its neighboring samples is calculated, then the similarity was combined with the actual subjection degree information to make a final judgement.
With the upgrading of detection instruments and other equipment, the ability to obtain information has improved, which may lead to the existence of some unknown classes of the elements out of the frame of discernment.
Hence, in reality, most of the frames of discernment may be in the open world.
In the open world, also named the incomplete frame of discernment,
Deng \cite{deng2015generalized} defined a novel concept called generalized basic probability assignment (GBPA) to model uncertain information, and provide a generalized combination rule (GCR) for the combination of GBPAs.
Zhang et al. \cite{zhang2017method} proposed a method using triangular fuzzy numbers to determine BPA in the open world, which can generate reasonable GBPA based on the prior sample data and decrease conflicts among the propositions.
Jiang et al. \cite{jiang2018improved} also put forward a new method to generate GBPA in the open world according to the triangular fuzzy numbers. It defined a pessimistic strategy based on the magnitude of the difference between samples and the model to obtain the known target's BPA.
However, most of the methods have the disadvantages of computational complexity, which may not meet the needs of real-time changes.
And the recognition accuracy of those method needs to be improved.
Consequently, how to effectively and quickly generate GBPA in the incomplete frame of discernment is still an open issue.

In the article \cite{deng2011methods}, Deng pointed out that the strongly constrained GBPA generation method can generate a GBPA with an empty set that is not equal to zero, and its size reflects the possibility that the system is the open world.
Zhang et al. \cite{zhang2017method} proposed a new method based on triangular fuzzy number to determine BPA in the open world.
In the article \cite{zhang2017method}, the mean value, standard deviation, extreme values as well as triangular membership function of each attribute are utilized to generate the fuzzy triangle curves, and construct the nested structure BPA function with an assignment for the null set.
With the in-depth study, Jiang et al. \cite{jiang2017modified} found that the GCR still has its issue. Therefore, a modified generalized combination rule (mGCR) in the framework of generalized evidence theory (GET) was proposed. The mGCR satisfies all properties of GCR in GET, illustrating and modeling the real world more reasonably than the original.
In comparison with the method in \cite{zhang2017method}, the contribution of the proposed method can be distinguished as follows:
\begin{enumerate}
  \item The proposed method simplifies the generation criterion of $m\left( {\left\{ \emptyset  \right\}} \right)$.
In  \cite{zhang2017method}, the sum of the generated GBPA values may exceed 1,
and the generation criterion of $m\left( {\left\{ \emptyset  \right\}} \right)$ depends not only on whether the sample intersects the triangular fuzzy numbers represented by propositions, but also on whether the sum of the GBPA values of other single-subset and multi-subset propositions is greater than 1.
In the proposed method, the sum of GBPA values for all propositions is less than or equal to 1,
and the generation criterion of $m\left( {\left\{ \emptyset  \right\}} \right)$ is just 1 minus the sum of GBPA values of all single-subset propositions and multi-subset propositions.
In this way, the generation criterion of $m\left( {\left\{ \emptyset  \right\}} \right)$ in this method is simpler.
  \item The proposed method improves the strategy for generation of GBPA.
When the sample intersects with the triangular fuzzy number representation model of multiple propositions, the strongly constrained GBPA generation strategy is: the high point of the ordinate is the GBPA that the sample supports the proposition, and the low point of the ordinate is the GBPA that the sample supports the multi-subset proposition.
The GBPA generation strategy in this paper is: the interval length of the high point of the ordinate minus the low point of the ordinate is the GBPA for which the sample supports a single or multiple subset proposition, the low point of the ordinate is the GBPA that the sample supports the multi-subset proposition.
  \item The method in this paper generalized the ability of modeling uncertain information in the open world by generating more non-zero values of $m\left( {\left\{ \emptyset  \right\}} \right)$.
 In \cite{zhang2017method}, $m\left( {\left\{ \emptyset  \right\}} \right)$ is often given 0 because the sum of non-empty sets' values of GBPA exceed 1, but in the proposed method, 0 will be assigned only when the sample and the triangle fuzzy number model of the proposition intersect the vertex.
Therefore, the improved method will have a better chance to generate $m\left( {\left\{ \emptyset  \right\}} \right)$ for incomplete information, which reduces the loss of external environmental information.
\end{enumerate}

The rest of the paper is arranged as follows. Section 2 introduces the preliminary  of this paper. Section 3 proposes a new method to generate GBPA. Section 4 uses UCI data sets to prove the effectiveness and superiority of the proposed method. Section 5 discusses the application of the proposed model in a closed world as well as the experiment basing on the improved Dempster combination rule. Section 6 draws a conclusion.

\section{Preliminaries}
\label{sec:2}
Some preliminaries are briefly introduced in this section, including Dempster-Shafer evidence theory \cite{shafer1976mathematical,dempster1967upper}, Generalized evidence theory \cite{deng2015generalized}, Triangle fuzzy number \cite{zadeh1965fuzzy,dubois1978operations}, and Modified generalized combination rule \cite{jiang2017modified}.

\subsection{Dempster-Shafer Evidence Theory}
\label{sec:2.1}
Dempster-Shafer evidence theory was proposed by Dempster \cite{dempster1967upper} and was then further developed by Shafer \cite{shafer1976mathematical}. In this part, a few concepts of Dempster-Shafer evidence theory are mainly introduced. The concepts are introduced as below.
\begin{definition}
Assume that $\Omega {\rm{ = }}\left\{ {{\theta _1},{\theta _2}, \ldots ,{\theta _i}, \ldots ,{\theta _N}} \right\}$ is a nonempty set with $N$ mutually exclusive and exhaustive events,  $\Omega$ is the frame of discernment (FOD). The power set of  $\Omega$ consists of ${2^N}$ elements denoted as follows:
\begin{equation}
{2^\Omega } = \left\{ \begin{array}{l}
 \emptyset ,\left\{ {{\theta _1}} \right\},\left\{ {{\theta _2}} \right\}, \ldots ,\left\{ {{\theta _N}} \right\},\left\{ {{\theta _1},{\theta _2}} \right\}, \\
  \ldots ,\left\{ {{\theta _1},{\theta _2}, \ldots ,{\theta _i}} \right\}, \ldots ,\Omega  \\
 \end{array} \right\}.
\end{equation}
\end{definition}

\begin{definition}
A mass function $m$ is a mapping from the power set ${2^\Omega }$ to the interval [0,1]. $m$ satisfies:
\begin{equation}
m\left( \emptyset  \right){\rm{ = }}0,
{}\sum\limits_{A \in \Omega } {m\left( A \right){\rm{ = }}1}.
\end{equation}
If $m\left( A \right) > 0$, then $A$ is called a focal element. $m\left( A \right)$ indicates the support degree of the evidence on the proposition $A$.
\end{definition}

\begin{definition}
The body of evidence (BOE), also known as basic probability assignment (BPA) or basic belief assignment (BBA), is defined as the focal sets and the corresponding mass functions:
\begin{equation}
\left( {\Re ,m} \right) = \left\{ {\left\langle {A,m\left( A \right)} \right\rangle :A \in {2^\Omega },m\left( A \right) > 0} \right\}.
\end{equation}
where $\Re $ is a subset of the power set ${{2^\Omega }}$.
\end{definition}
\begin{definition}
A BPA $m$ can also be represented by the belief function $Bel$  or the plausibility function  $Pl$ , defined as follows:
\begin{equation}
Bel\left( A \right){\rm{ = }}\sum\limits_{\emptyset  \ne B \subseteq A} {m\left( B \right)} \begin{array}{*{20}{c}}
{}
\end{array}, \begin{array}{*{20}{c}}
{}
\end{array}Pl\left( A \right){\rm{ = }}\sum\limits_{B \cap A \ne \emptyset } {m\left( B \right)} .
\end{equation}
\end{definition}

\begin{definition}
In Dempster-Shafer evidence theory, 2 independent mass functions ${m_1}$ and ${m_2}$ can be fused with Dempster's rule of combination:
\begin{equation} \label{eq5}
m(A){\rm{ = }}\left( {m_1  \oplus m_2 } \right)\left( A \right){\kern 1pt} {\rm{ = }}\frac{1}{{1 - k}}\sum\limits_{B \cap C = A} {m_1 (B)  m_2 (C)},
\end{equation}
where ${k}$ is a normalization factor defined as follows:
\begin{equation}
k{\rm{ = }}\sum\limits_{B \cap C = \emptyset } {m_1 (B)  m_2 (C)}.
\end{equation}
\end{definition}

\subsection{Generalized evidence theory}
\label{sec:2.2}
The basic concepts of generalized evidence theory (GET) is introduced in this section \cite{deng2015generalized}.
In GET, the generalized basic probability assignment (GBPA) corresponds to BPA in Dempster-Shafer evidence theory, which is used for data expression and modeling, and the generalized combination rule (GCR) is provided for combining conflicting or inconsistent or uncertain evidence.

\begin{definition}
Suppose that $U$ is a frame of discernment in an open world. Its power set, $2_G^U $ , is composed of ${2^U}$ propositions, $\forall A \subset U$.
A mass function is a mapping ${m_G}:2_G^U \to [0,1]$, that satisfies
\begin{equation}\label{GET1}
\sum\limits_{A \in 2_G^U} {{m_G}\left( A \right) = 1.}
\end{equation}
\end{definition}
Then, ${m_G}$  is the GBPA of the frame of discernment, $U$.
The difference between GBPA and traditional BPA is the restriction of $\emptyset $. In GET, $ {m_G}\left( \emptyset  \right){\rm{ = }}0 $ is not necessary in GBPA, and this is a big difference from conventional Dempster-Shafer evidence theory.
In other words, the empty set can also be a focal element. If $ {m_G}\left( \emptyset  \right){\rm{ = }}0 $, the GBPA reduces to a conventional BPA in Dempster-Shafer evidence theory.

\begin{definition}
Given a GBPA $m$, the GBF is GBel: ${2^U} \to [0,1]$, which satisfies
\begin{equation}\label{GET1.1}
\begin{array}{*{20}{c}}{GBel(A)}& = &{\sum\limits_{B \subseteq A} {m(B)} ,}\end{array}
\end{equation}
and
\begin{equation}\label{GET1.11}
\begin{array}{*{20}{c}}{GBel(\emptyset  )}& = &{m(\emptyset ).}\end{array}
\end{equation}
\end{definition}

\begin{definition}\label{def2.3}
Given a GBPA $m$, the GBF is GPl: ${2^U} \to [0,1]$, which satisfies
\begin{equation}\label{GET1.2}
{GPl(A) = \sum\limits_{B \cap A \ne \emptyset } {m(B)}, }
\end{equation}
and
\begin{equation}\label{GET1.21}
\begin{array}{*{20}{c}} {GPl(\emptyset ) = m(\emptyset ).} \end{array}
\end{equation}
\end{definition}

\begin{definition}\label{def2.4}
In generalized evidence theory, ${\emptyset _1} \cap {\emptyset _2} = \emptyset $ means that the intersection between two empty sets is still an empty set. Given two GBPAs (${m_1}$ and ${m_2}$), the GCR is defined as follows.
\begin{equation}\label{GET2}
m\left( A \right) = \frac{{\left( {1 - m\left( \emptyset  \right)} \right)\sum\limits_{B \cap C = A} {{m_1}\left( B \right) \cdot {m_2}\left( C \right)} }}{{1 - K}},
\end{equation}

\begin{equation}\label{GET2.1}
\begin{array}{*{20}{c}} K& = &{\sum\limits_{B \cap C = \emptyset } {{m_1}(B) \cdot {m_2}(C)} ,} \end{array}
\end{equation}

\begin{equation}\label{GET2.2}
\begin{array}{*{20}{c}} {m(\emptyset )}& = &{{m_1}(\emptyset ) \cdot {m_2}(\emptyset ),} \end{array}
\end{equation}

\begin{equation}\label{GET2.3}
\begin{array}{*{20}{c}} {m(\emptyset )}& = &{1\quad if\;\;and\;\;only\;\;if\quad K = 1.} \end{array}
\end{equation}
\end{definition}

\begin{definition}
Jiang et al. proposed a modified generalized combination rule (mGCR) \cite{jiang2017modified} to improve the shortcomings of the generalized evidence combination rule.
Given two sets of GBPAs $\left( {{m_1},{m_2}} \right)$, the modified generalized combination rule is defined as follows:
\begin{equation}
\begin{array}{l}
m\left( A \right) = \frac{{\sum\limits_{B \cap C = A} {{m_1}\left( B \right)\cdot{m_2}\left( C \right)} }}{{1 - K}},\\
\end{array}
\end{equation}
\begin{equation}
\begin{array}{l}
m\left( \emptyset  \right) = \frac{{{m_1}\left( \emptyset  \right)\cdot{m_2}\left( \emptyset  \right)}}{{1 - K}},
\end{array}
\end{equation}
with
\begin{equation}\label{GET2.1}
\begin{array}{*{20}{c}} K& = &{\sum\limits_{B \cap C = \emptyset ,B \cup C \ne \emptyset } {{m_1}\left( \emptyset  \right) \cdot {m_2}\left( \emptyset  \right)} ,} \end{array}
\end{equation}
\begin{equation}
\begin{array}{l}
m\left( \emptyset  \right) = 1 \quad if\;\;and\;\;only\;\;if\; K = 1 \quad or\;\sum\limits_{A \ne \emptyset } {m\left( A \right) = 0.}
\end{array}
\end{equation}
\end{definition}

\subsection{Triangle Fuzzy Number}
\label{sec:2.3}
The theory of fuzzy number \cite{dubois1978operations} is based on the fuzzy set which is proposed by Zadeh in 1965 \cite{zadeh1965fuzzy}, a fuzzy set in a certain domain of discourse $U$ can be represented as a membership function $f\left( x \right)$. For $\forall x \in U$ there always exists a $f\left( x \right) \in [0,1]$. And $f\left( x \right)$ is called the membership degree of $x$ to $U$.

A triangular fuzzy number $\left( {{a_0},{\mu _\alpha },{a_1}} \right)$ is a fuzzy number with a piecewise liner membership function $f\left( x \right)$ defined by:
\begin{equation}
f\left( x \right) = \left\{ {\begin{array}{*{20}{r}}
0&{x < {a_0}}\\
{\frac{{x - {a_0}}}{{{\mu _\alpha } - {a_0}}}}&{{a_0} \le x \le {\mu _\alpha }}\\
{\frac{{{a_1} - x}}{{{a_1} - {\mu _\alpha }}}}&{{\mu _\alpha } \le x \le {a_1}}\\
0&{x > {a_1}}
\end{array}} \right..
\end{equation}
A fuzzy number is limited with its upper bound ${{a_1}}$ and lower bound ${{a_0}}$. Symbol ${{\mu _\alpha }}$ corresponds the most likely value.

\section{A new approach for the generation of GBPA}
\label{sec:3}
\label{sec:3}
Based on the multi-subset propositional representation model of fuzzy mathematics, this paper proposes a method to generate GBPA in the incomplete framework of discernment.
The prerequisite is to have prior knowledge of the maximum, minimum, and mean values in the sample data of all classes.
Hence, the proposed method is suitable for practical projects that contain an amount of sample data.
The idea of the proposed method is that if a sample is between the maximum and minimum values of multiple classes, it is considered that the sample has GBPA that supports multiple propositions and GBPA that supports the single proposition.
The detailed steps of the proposed method are as follows:
\paragraph{Step 1} Set the minimum of the triangular membership function to 0.01.
When the attribute value of the test sample is equal to the maximum or minimum of the triangular fuzzy number model, it is unreasonable to assign the membership degree of the attribute to 0.
Therefore, the minimum of the triangle membership function is set to 0.01, so that it can avoid unreasonable situations and not change the boundary value of the triangle fuzzy number model too much.

\paragraph{Step 2}When the test sample intersects with the triangular fuzzy number models, $m$ intersection points will be produced. Symbolize the ordinate values in ascending order as ${w_1},{w_2}, \cdots {w_m}$.

\begin{figure}
  \centering
  \includegraphics[scale=0.5]{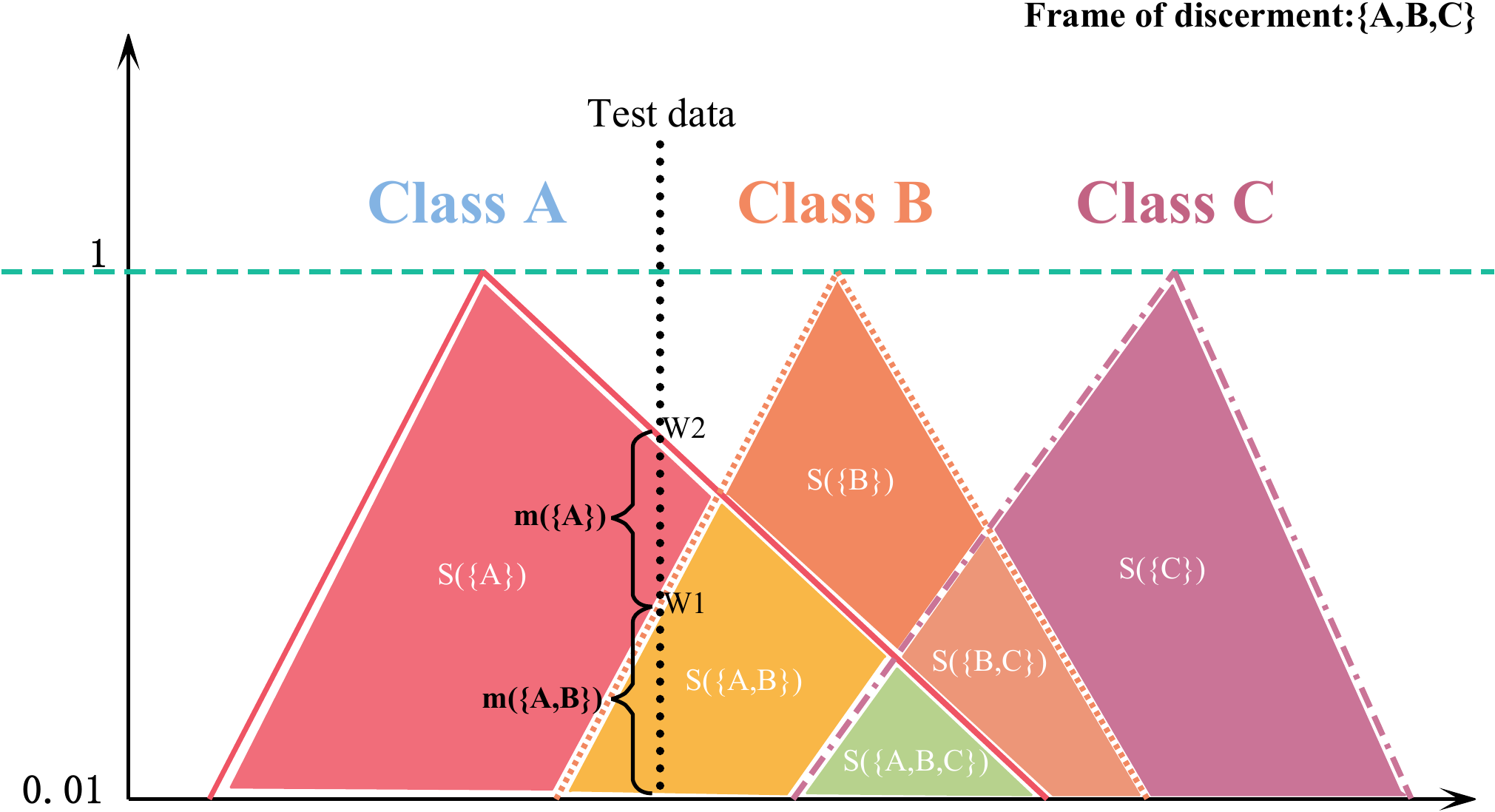}\\
  \caption{The generation of GBPA}\label{fig:generate}
\end{figure}

\paragraph{Step 2.1} Determine the interval of ${w_m}$ to ${w_{m - 1}}$ belongs to which triangular fuzzy number model that contains the most propositional subsets.
\paragraph{Step 2.2}The ordinate of ${w_m}$ minus the ordinate of ${w_{m - 1}}$ is the GBPA for which multi-subset proposition the sample supports.
Let $m\left( A \right)$ represent the BPA value of the focal element $A$. Figure \ref{fig:generate} is the visual form of this rule. The detailed rule are as follows:
\begin{equation}
\begin{array}{l}
m\left( {\left\{ {{C_1},{C_2}, \cdots {C_m}} \right\}} \right) = {w_1}\\
m\left( {\left\{ {{C_2}, \cdots {C_m}} \right\}} \right) = {w_2} - {w_1}\\
 \cdots \\
m\left( {\left\{ {{C_{m - 1}},{C_m}} \right\}} \right) = {w_{m - 1}} - {w_{m - 2}}\\
m\left( {\left\{ {{C_m}} \right\}} \right) = {w_m} - {w_{m - 1}}
\end{array}.
\end{equation}

\paragraph{Step 3}Assign the value that 1 minus the sum of generated GBPA values to $m\left( {\left\{ \emptyset  \right\}} \right)$.    Because the framework of discernment is incomplete, $m\left( {\left\{ \emptyset  \right\}} \right)$ is used to represent the influence of external environmental factors.

\begin{figure}
  \centering
  \includegraphics[scale=0.5]{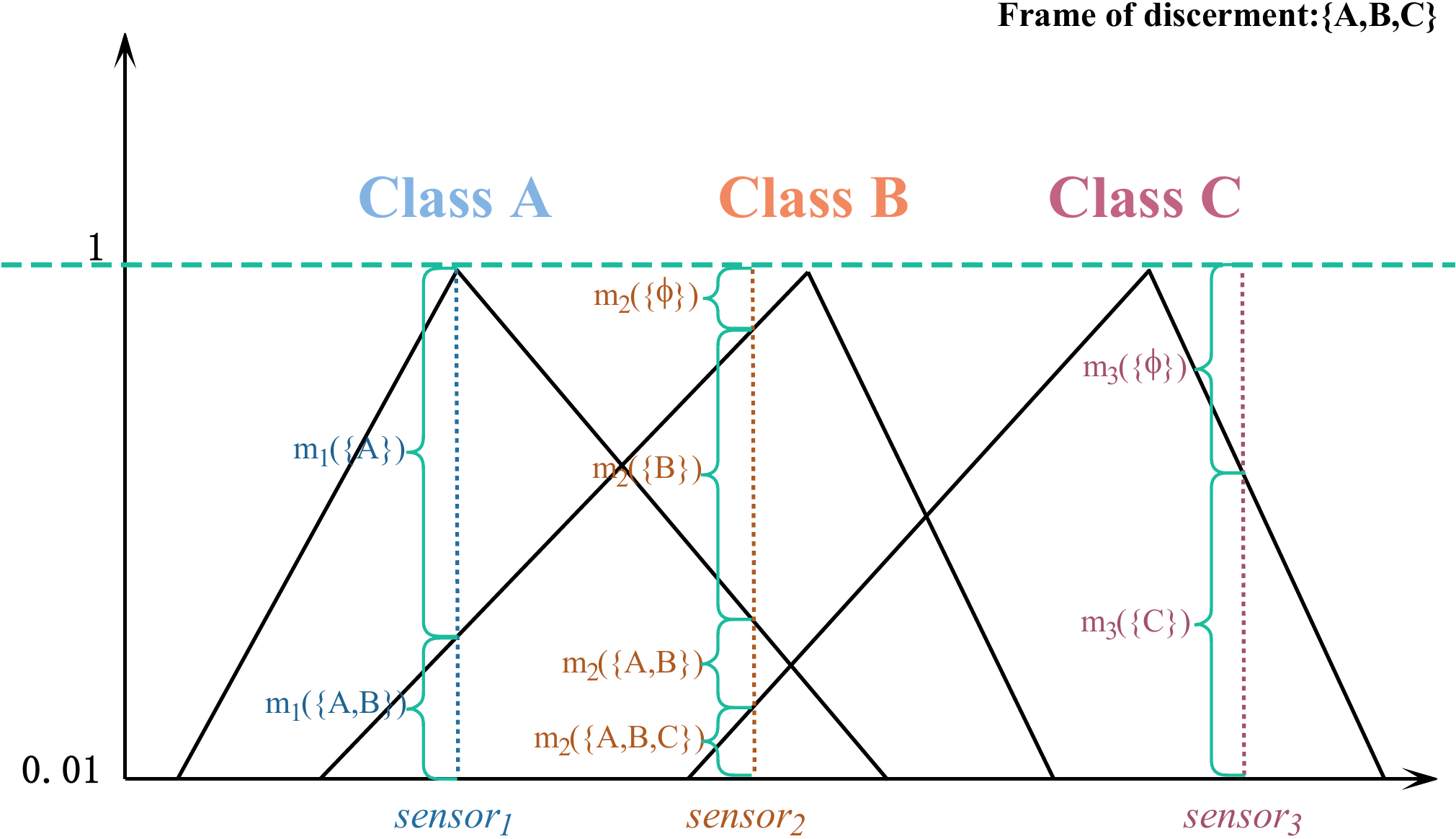}\\
  \caption{The example of generation of GBPA}\label{fig:example}
\end{figure}

In Figure \ref{fig:example}, the ordinate represents the GBPA value, and the abscissa represents the data detected by the sensor.
For sensor 3, the test sample only intersects the triangle fuzzy number model of class C.
Obviously, the test sample's GBPAs is divided into 2 parts. The lower part of the interval is in the triangle fuzzy number model of class C, so the length of the interval is the value of ${m_3}\left( {\left\{ C \right\}} \right)$. The upper part of the interval is not in any triangle fuzzy number models, so the length value of this interval is assigned to ${m_3}\left( {\left\{ \emptyset  \right\}} \right)$.
For sensor 2, the test sample intersects the triangle fuzzy number model of class A, class B and class C. The test sample's GBPA is divided into 4 parts.
The bottom interval is in the common representation model of class A, class B and class C, so its length is equal to ${m_2}\left( {\left\{ {A,B,C} \right\}} \right)$,
and the interval above it is in the representation model of class A and class B, so the length of the interval is equal to ${m_2}\left( {\left\{ {A,B} \right\}} \right)$ and so on. Through the above steps, the generation of GBPA for the second sensor can be completed.

\section{Experiment}
\label{sec:4}
In this section, a series of experiments for verification of the proposed method is designed.
These experiments not only use the $Haberman's Survival$ data set to demonstrate the generation of GBPA, but also use $Iris$ data set to test the rationality and performance of the proposed method, as well as analyze the robustness of classification accuracy.

\subsection{An example to generation of GBPA}
\label{sec:4.1}
In this section, $Haberman's Survival$ data set is used to demonstrate the effectiveness of GBPA generation in the open world.
The data set contains cases from a study that was conducted between 1958 and 1970 at the University of Chicago's Billings Hospital.
The study counted the age of patients at time of operation, the patients' year of operation, the number of positive axillary nodes detected, and the patients' final survival status. Of these cases, 225 patients survived 5 years or longer, and 81 patients died within 5 years.

\begin{figure}
  \centering
  \includegraphics[scale=0.33]{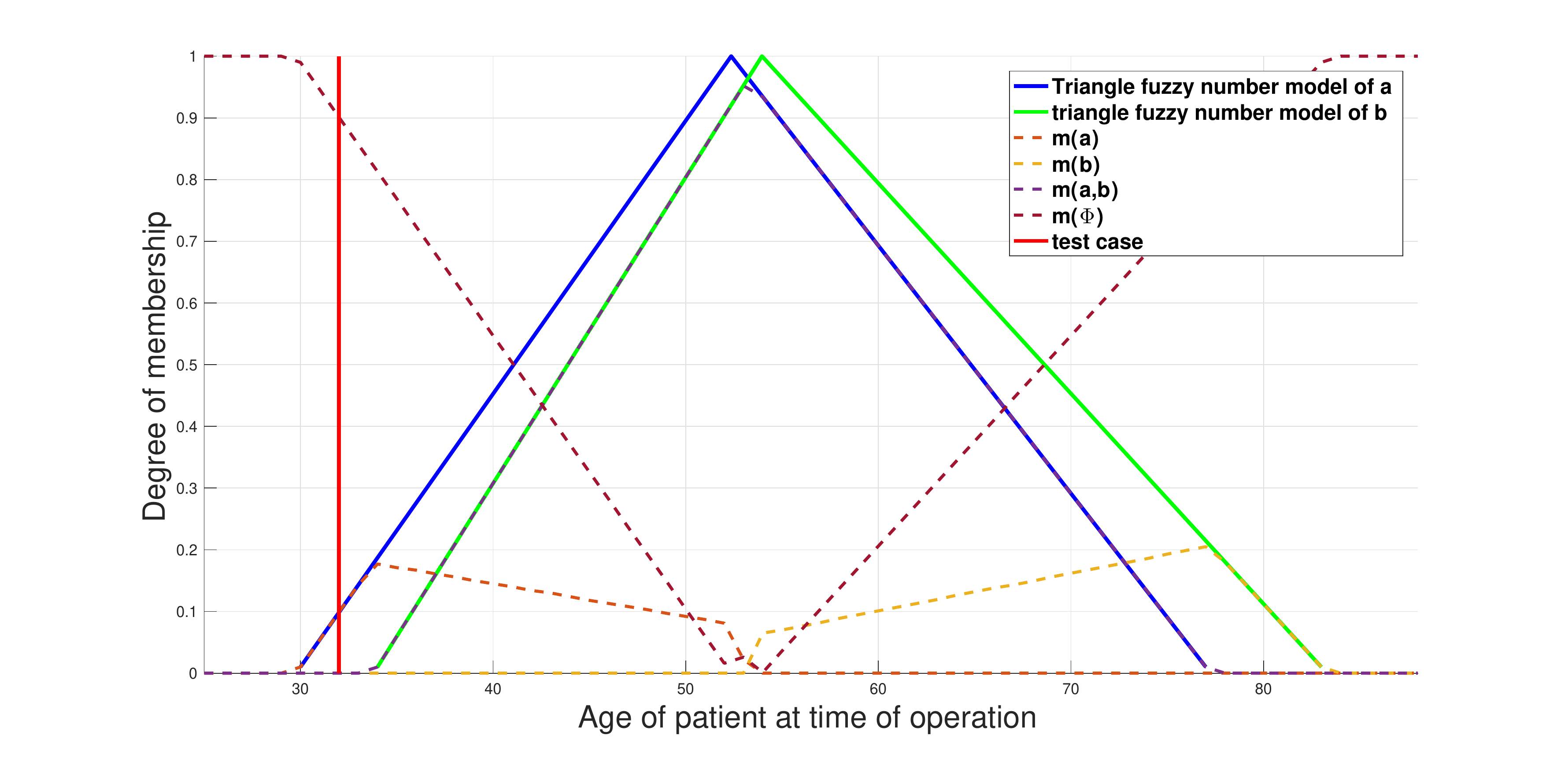}\\
  \caption{Curves in attribute of the age of patients at time of operation}\label{fig:generate1}
\end{figure}

\begin{figure}
  \centering
  \includegraphics[scale=0.33]{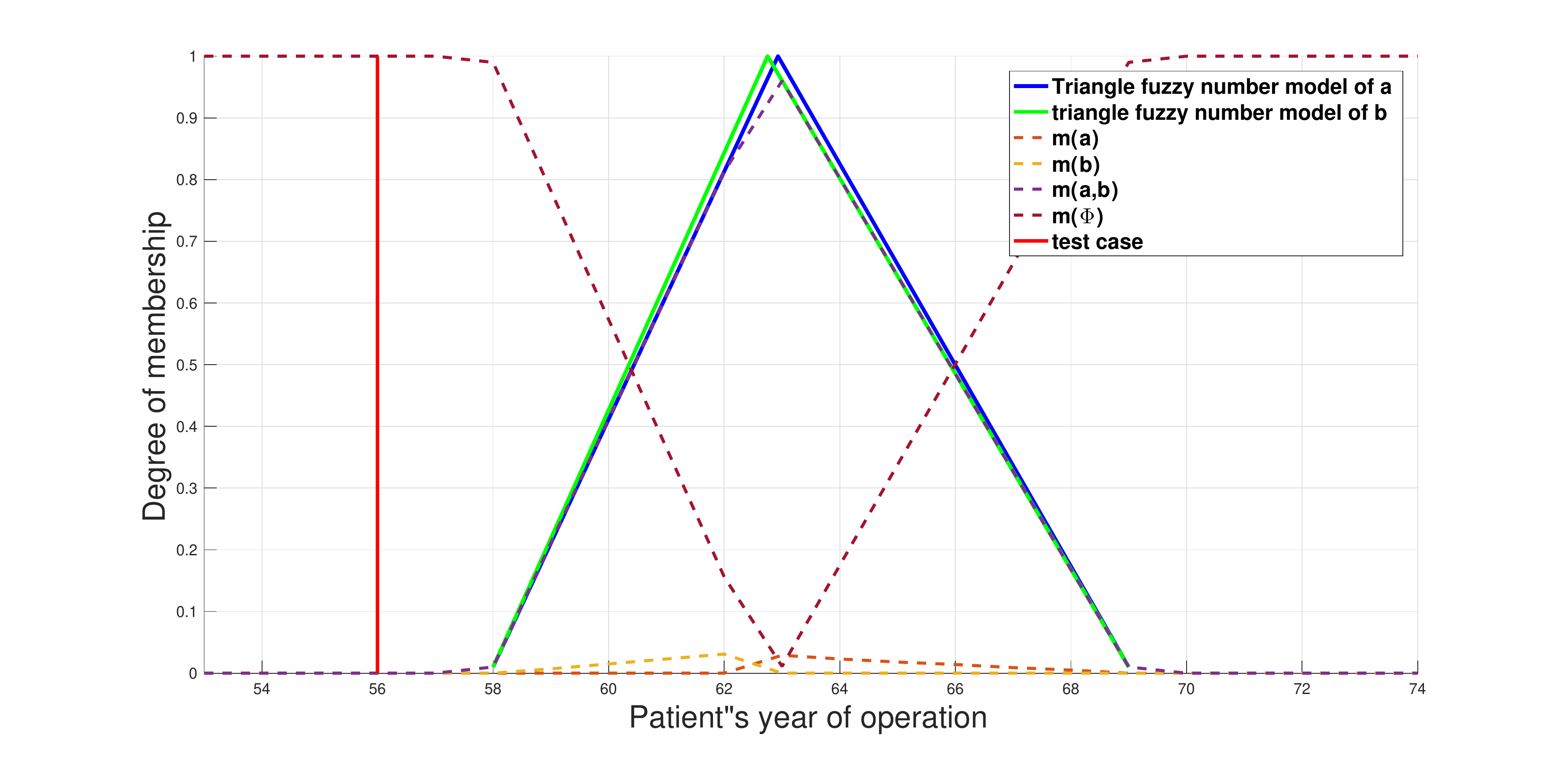}\\
  \caption{Curves in attribute of the patients' year of operation}\label{fig:generate2}
\end{figure}

\begin{figure}
  \centering
  \includegraphics[scale=0.33]{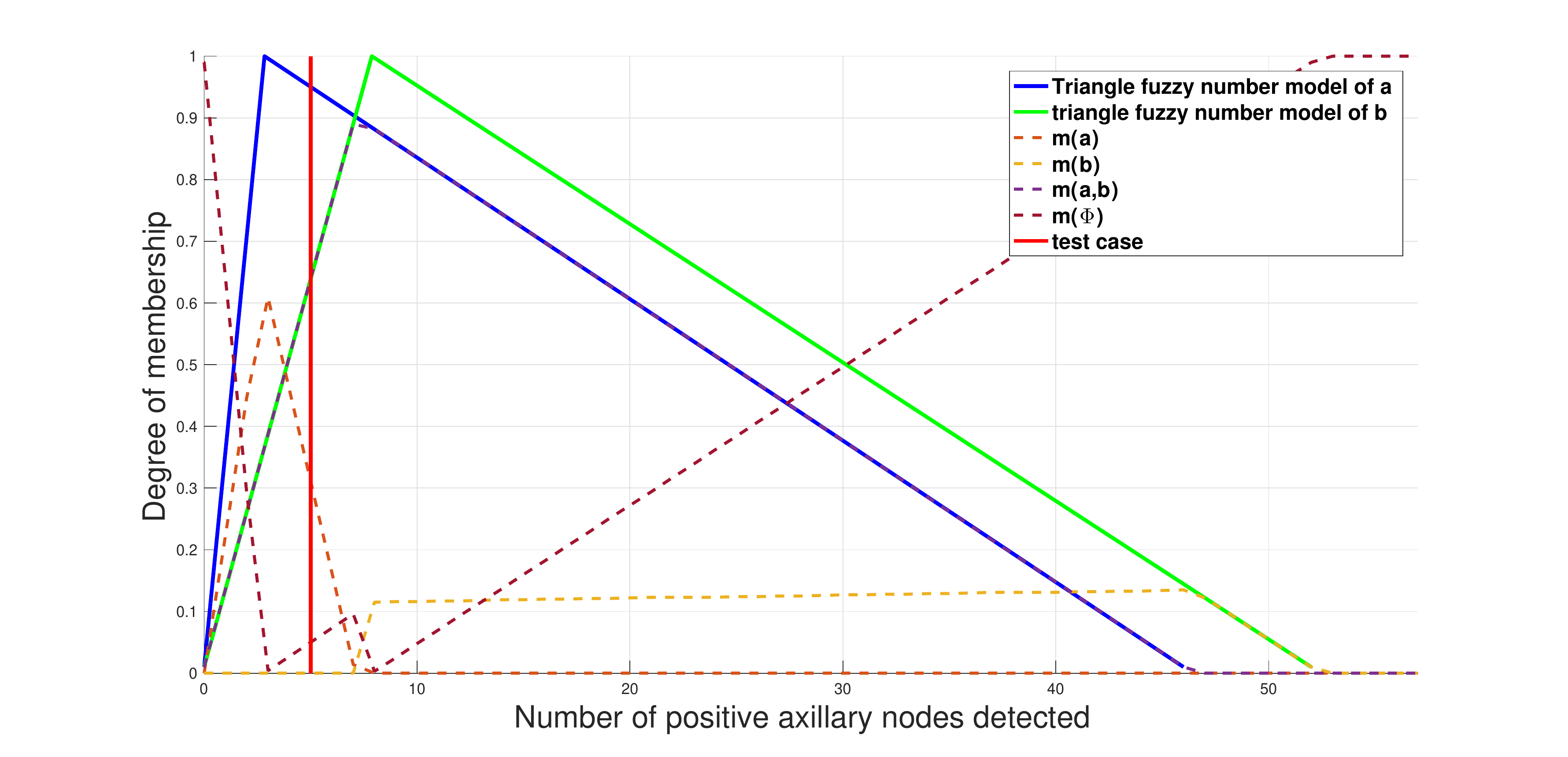}\\
  \caption{Curves in attribute of the number of positive axillary nodes detected}\label{fig:generate3}
\end{figure}

In the $Haberman's Survival$ data set, the framework of discernment is set to $\Theta  = \left\{ {a,b} \right\}$, event $a$ represents the result of the patient's survival for 5 years or more, and the event $b$ represents the result of the patient's death within 5 years.
In events $a$ and $b$, 200 and 72 cases are randomly selected as the training set, and the remaining samples are used as the test set.
In the proposed method, each of
the four attributes is treated as an independent information source, and test samples' GBPA are generated with a step length of 1. Each GBPA of the test samples is marked on the triangular fuzzy number representation models.
The distribution of GBPA on the frame of discernment of the age of patients at time of operation is shown in Figure \ref{fig:generate1}.
The distribution of GBPA on the frame of discernment of  the patients' year of operation is shown in Figure \ref{fig:generate2}.
The distribution of GBPA on the frame of discernment of  the number of positive axillary nodes detected is shown in Figure \ref{fig:generate3}.

Figure \ref{fig:generate1}, Figure \ref{fig:generate2}, and Figure \ref{fig:generate3} show the following features. 
\paragraph{1)} When the test sample is not in the interval of the triangular fuzzy number model of class a and class b, as shown in Figure \ref{fig:generate2} where the sample's value is less than 58, the test sample has no intersection with the triangular fuzzy number model of class a and class b.
Therefore, it is easy to get $m\left( {\{ \emptyset \} } \right) = 1$, and $m\left( {\{ a\} } \right) = 0$, $m\left( {\{ b\} } \right) = 0$, which means that the system cannot determine which target the test sample belongs to, or which target it is more inclined to belong to.
In practical engineering applications, when class a and class b are not within the frame of discernment, it is a reasonable behavior to consider the test sample as an unknown class.
\paragraph{2)} When the test sample is in the triangular fuzzy number of class a but not in the triangular fuzzy number of class b, as shown in Figure \ref{fig:generate1} where the test sample's value is equal to 32, only $m\left( {\{ \emptyset \} } \right)$ and $m\left( {\{ a\} } \right)$ are available for GBPA values greater than 0, and $m\left( {\{ \emptyset \} } \right)$ decreases as $m\left( {\{ a\} } \right)$ increases in the frame of discernment.
If the system supports class $a$ more and more, it indicates that the system is more and more convinced that the discriminated class belongs to class a, rather than unknown classes.
\paragraph{3)}When the test sample is in the mixed triangular fuzzy number of class a and  class b, a nested set is generated.
The value of $m\left( {\{ a\} } \right)$ and $m\left( {\{ b\} } \right)$ decreases or does not change as the value of $m\left( {\{ a,b\} } \right)$ increases.
The value of $m\left( {\{ a,b\} } \right)$ reaches the maximum when the test sample is at the intersection point of the two triangular fuzzy number representation models of class a and class b. As shown in Figure \ref{fig:generate3} where the number of positive axillary nodes detected is equal to 7, the system considers that the recognized object is most likely to belong to $\left\{ {a,b} \right\}$. In the common interval of triangular fuzzy number models class a and class b, the value of $m\left( {\left\{ \emptyset  \right\}} \right)$ is related to $m\left( {\left\{ a \right\}} \right)$, $m\left( {\left\{ b \right\}} \right)$, $m\left( {\left\{ a,b \right\}} \right)$, and their relationship can be expressed as $m\left( {\left\{ \emptyset  \right\}} \right) = 1 - m\left( {\left\{ a \right\}} \right) - m\left( {\left\{ b \right\}} \right) - m\left( {\left\{ {a,b} \right\}} \right)$.

\subsection{Application in classification of Iris data set in the open world}
\label{sec:4.2}
In this section, data set of $Iris$ $D$ is used to verify the rationality and performance of the proposed method.
In the $Iris$ data set, there are three classes of iris, namely Setosa (a), Versicolor (b) and Virginia (c), each of which contains 50 samples. Each sample includes four attributes, namely calax length (SL), calax width (SW), petal length (PL), and petal width (PW).
40 samples are randomly selected from each class as the training set, use the maximum, minimum, and mean values of each attribute of each class in the training set to construct the corresponding triangular fuzzy number model, and set the remaining samples as the test set to verify the rationality of the proposed method.
The following tables show the process of using a test sample from class a, generating GBPAs and using the modified generalized combination rule for data fusion in the open world.

We calculate the maximum, minimum, and mean values of the four attributes of the three iris classes in the training set, so that we can get the triangular fuzzy number of any attribute in each iris class. The detailed data of these 12 groups of triangular fuzzy numbers are shown in Table \ref{tab:numbers}.

\begin{table}[htbp]
  \centering
  \caption{Triangular fuzzy numbers from the training set}
  \resizebox{\textwidth}{9mm}{ 
    \begin{tabular}{lcccc}
    \toprule
    Class      & SL    & SW    & PL    & PW \\
    \midrule
    a     & (4.3000, 4.9975,5.8000) & (2.3000,3.4125,4.2000) & (1.0000,1.4650,1.9000) & (0.1000,0.2525,0.6000) \\
    b     & (4.9000, 5.9775,7.0000) & (2.0000,2.7750,3.4000) & (3.0000,4.2425,5.0000) & (1.0000,1.3275,1.8000) \\
    c     & (4.9000, 6.5050,7.7000) & (2.2000,2.9600,3.8000) & (4.5000,5.4850,6.9000) & (1.5000,2.0150,2.5000) \\
    \bottomrule
    \end{tabular}}%
  \label{tab:numbers}%
\end{table}%

A test sample was randomly selected from class a to generate GBPA. The test sample's attribute values on SL, SW, PL, and PW are shown in Table \ref{tab:sample}.

\begin{table}[htbp]
  \centering
  \caption{The attributes' value of the sample}
  \setlength{\tabcolsep}{6mm}{
    \begin{tabular}{ccccc}
    \toprule
  Attribute&  SL    & SW    & PL    & PW \\
    \midrule
   Value &  5.1   & 3.8   & 1.5   & 0.3 \\
    \bottomrule
    \end{tabular}}%
  \label{tab:sample}%
\end{table}%

\begin{figure}
  \centering
  \includegraphics[scale=0.36]{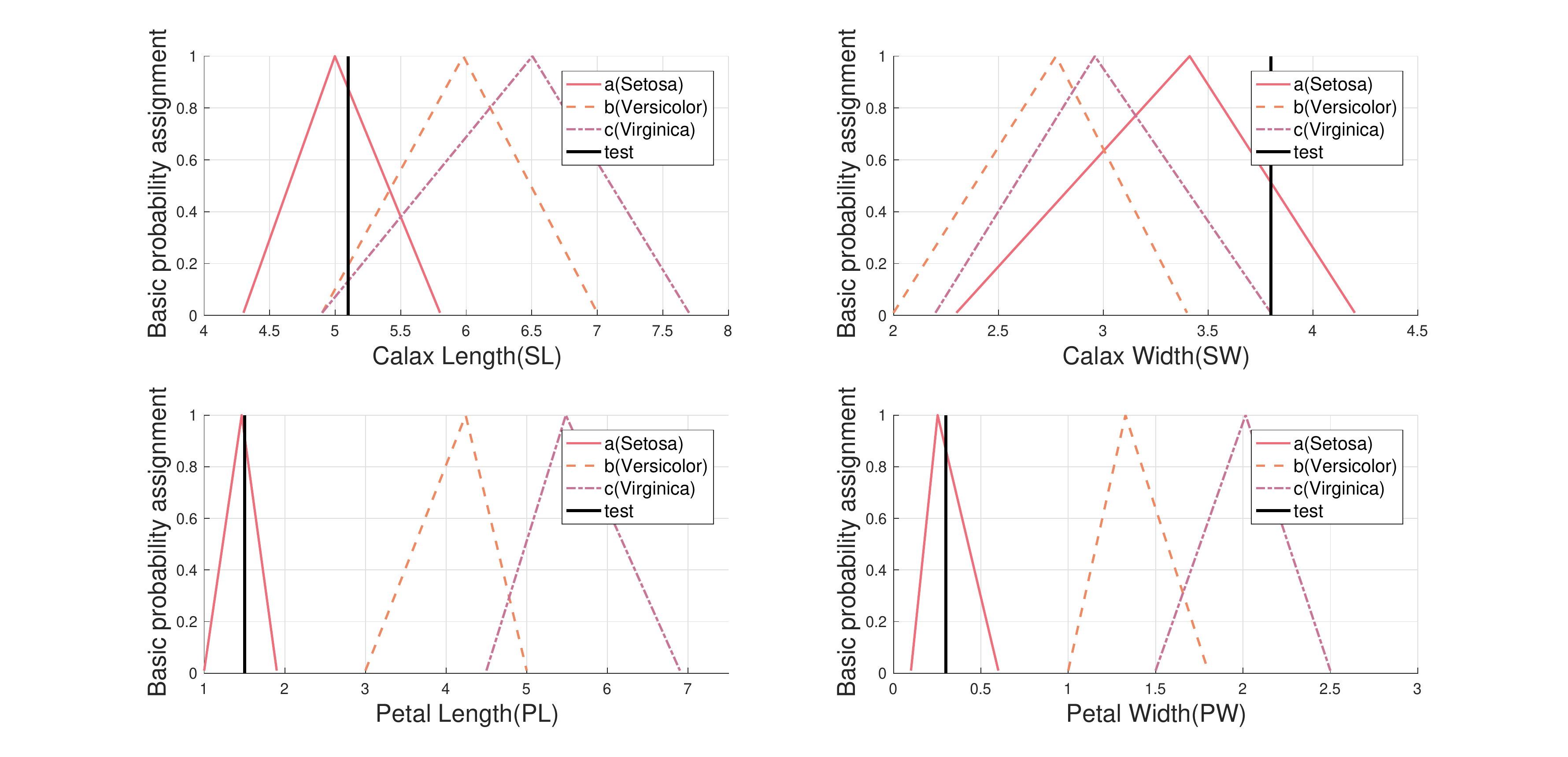}\\
  \caption{Intersection of test samples with triangular fuzzy number model}\label{fig:triangular}
\end{figure}

During classification, the attribute of the test sample is considered as independent information sources.
In this way, driven by a certain attribute value of the test sample, GBPAs are obtained according to the intersection of the attribute value and the triangular fuzzy number models.
The intersection of the four attribute values of the test sample with triangular fuzzy number models of a, b, and c is shown in Figure \ref{fig:triangular}.

According to the proposed in this paper, the SL, SW, PL, and PW attribute values of the test samples are assigned according to the intersection with the triangular fuzzy number models. The assigned results are shown in Table \ref{tab:GBPAs}.
When the attribute value is at the boundary of one triangular fuzzy number model, as shown in Figure \ref{fig:triangular} where the value of SW is at the maximum of the triangular fuzzy number model of class b, its membership is 0.01, not 0.

\begin{table}[htbp]
  \centering
  \caption{GBPAs generated from test sample in the open world}
  \resizebox{\textwidth}{10mm}{
    \begin{tabular}{ccccccccc}
    \toprule
        Attribute   &$m\left( {\left\{ a \right\}} \right)$      &$m\left( {\left\{ b \right\}} \right)$       &$m\left( {\left\{ c \right\}} \right)$       &$m\left( {\left\{ a,b \right\}} \right)$       &$m\left( {\left\{ a,c \right\}} \right)$       &$m\left( {\left\{ b,c \right\}} \right)$       &$m\left( {\left\{ a,b,c \right\}} \right)$       &$m\left( {\left\{ \emptyset  \right\}} \right)$  \\
    \midrule
    SL    & 0.680  & 0     & 0     & 0.061 & 0     & 0     & 0.133 & 0.126 \\
    SW    & 0.503 & 0     & 0     & 0     & 0.010  & 0     & 0     & 0.487 \\
    PL    & 0.920  & 0     & 0     & 0     & 0     & 0     & 0     & 0.080 \\
    PW    & 0.865 & 0     & 0     & 0     & 0     & 0     & 0     & 0.135 \\
    \bottomrule
    \end{tabular}}%
  \label{tab:GBPAs}%
\end{table}%
After using the proposed method to generate GBPAs for each information source, mGCR was used to fuse the GBPAs generated by the four groups of information sources in Table \ref{tab:GBPAs}. The final fusion results are shown in Table \ref{tab:classification}.
\begin{table}[htbp]
  \centering
  \caption{the fusion result of the test sample in the open world}
  \resizebox{\textwidth}{6mm}{
    \begin{tabular}{ccccccccc}
    \toprule
        BPA  & $m\left( {\left\{ a \right\}} \right)$      & $m\left( {\left\{ b \right\}} \right)$      & $m\left( {\left\{ c \right\}} \right)$      & $m\left( {\left\{ a,b \right\}} \right)$      & $m\left( {\left\{ a,c \right\}} \right)$      & $m\left( {\left\{ b,c \right\}} \right)$    & $m\left( {\left\{ a,b,c \right\}} \right)$      & $m\left( {\left\{ \emptyset  \right\}} \right)$    \\
    \midrule
   Value& 0.9981 & 0     & 0     & 0     & 0     & 0     & 0     & 0.0019 \\
    \bottomrule
    \end{tabular}}%
  \label{tab:classification}%
\end{table}%
It can be seen from the fusion results that the proposed method has a good performance.
In the open world with environmental interference, it still assigns a very high belief value to the correct class a, which means a good recognition effect.
The experiment in this section reflects the rationality and easy recognition of the proposed method.

\subsection{Classification testing in the open world}
\label{sec:4.3}
In this section, the frame of discernment of the data set of $Iris$ is divided into three cases to model respectively:
\begin{equation}
{\Theta _1} = \left\{ {a,b} \right\},{\Theta _2} = \left\{ {a,c} \right\},{\Theta _3} = \left\{ {b,c} \right\},
\end{equation}
where $a$ stands for class Setosa, $b$ stands for class Versicolor, and $c$ stands for class Virginica.
For example, in the framework of discernment ${\Theta _1}$, for each test sample, four sets of GBPAs will be generated according to their attributes.
If the actual class of the test sample is $a$, then the maximum belief value should be assigned to $m\left( {\left\{ a \right\}} \right)$.
Since the class $b$ is not in the framework of discernment, for the test sample whose actual class is $b$, a high belief value assigned to $m\left( {\left\{ \emptyset  \right\}} \right)$ is reasonable.

The result of the proposed method with the true class of test samples is compared in three frameworks of discernment respectively.
In each framework of discernment, 40 samples are randomly selected from each class as the training set, and the remaining 10 samples are used as the test set. In addition, 10 samples are randomly selected from the "unknown class" and added to the test set to serve as unknown factors in the open world.
Each attribute of the $Iris$ data set is still considered as an independent information source.
After using mGCR to fuse all evidences of the test sample, a new BOE is generated.
The final decision rule in the new BOE is to select the class corresponding to the maximum GBPA.
The results of three frameworks of discernment are shown in Table \ref{tab:frame_ab}, Table \ref{tab:frame_ac}, and Table \ref{tab:frame_bc}. The accuracy of the final result reflects the high efficiency of the proposed method.
\begin{table}[htbp]
  \centering
  \caption{Classification results in the open world ${\Theta _1} = \left\{ {a,b} \right\}$}
  \resizebox{\textwidth}{11mm}{
    \begin{tabular}{cccccc}
    \toprule
    Sample & Actual class & Ideal class & Sample number & Correct number  & Accuracy \\
    \midrule
       $sampl{e_1}$   & a     & a     & 20    & 16    & 80.00\% \\
       $sampl{e_2}$   & b     & b     & 20    & 20    & 100.00\% \\
       $sampl{e_3}$   & c     & $\emptyset$       & 20    & 20    & 100.00\% \\
    Total & -     & -     & 60    & 56    & 93.30\% \\
    \bottomrule
    \end{tabular}}%
  \label{tab:frame_ab}%
\end{table}%

\begin{table}[htbp]
  \centering
  \caption{Classification results in the open world ${\Theta _2} = \left\{ {a,c} \right\}$}
  \resizebox{\textwidth}{11mm}{
    \begin{tabular}{cccccc}
    \toprule
    Sample & Actual class & Ideal class & Sample number & Correct number  & Accuracy \\
    \midrule
          $sampl{e_1}$   & a     & a     & 20    & 20    & 100\% \\
          $sampl{e_2}$   & b     & $\emptyset$        & 20    & 20    & 100\% \\
          $sampl{e_3}$   & c     & c     & 20    & 20    & 100\% \\
    Total & -     & -     & 60    & 60    & 100\% \\
    \bottomrule
    \end{tabular}}%
  \label{tab:frame_ac}%
\end{table}%

\begin{table}[htbp]
  \centering
  \caption{Classification results in the open world ${\Theta _3} = \left\{ {b,c} \right\}$}
  \resizebox{\textwidth}{11mm}{
    \begin{tabular}{cccccc}
    \toprule
    Sample & Actual class & Ideal class & Sample number & Correct number  & Accuracy  \\
    \midrule
          $sampl{e_1}$   & a     & $\emptyset$        & 20    & 20    & 100.00\% \\
          $sampl{e_2}$   & b     & b     & 20    & 20    & 100.00\% \\
          $sampl{e_3}$   & c     & c     & 20    & 15    & 75.00\% \\
    Total & -     & -     & 60    & 55    & 91.60\% \\
    \bottomrule
    \end{tabular}}%
  \label{tab:frame_bc}%
\end{table}%

\section{Discussion}
\label{sec:5}
A good decision usually has downward compatibility.
When the whole set of BPA is assigned to be singleton propositions in the frame of discernment, the BPA is equal to the probability in probability theory.
As another example, if a system is in the closed world, the GCR assigns $m\left( {\left\{ \emptyset  \right\}} \right)$ to 0, and then the GCR degenerates to the Dempster combination rule. Similarly, the proposed method can also be used in the complete frame of discernment. Only the final step of the proposed method is needed to change. Let the value of 1 minus the sum of GBPAs value be evenly distributed to the other ${2^N} - 1$ BPAs ($N$ is the cardinality of the frame of discernment), indicating that each event is equally likely to occur.
\subsection{Compatibility in classification of Iris data set in the closed world}
\label{sec:5.1}
The value of $m\left( {\left\{ \emptyset  \right\}} \right)$ is evenly distributed in Table \ref{tab:GBPAs} to other BPAs, so that the BPA is constructed in the closed world. The distribution results in the closed world are shown in Table \ref{tab:BPAs}, where the fraction represents the value evenly obtained from the value of $m\left( {\left\{ \emptyset  \right\}} \right)$.

\begin{table}[htbp]
  \centering
  \caption{BPA generated from the test sample in the closed world}
  \resizebox{\textwidth}{12mm}{
    \begin{tabular}{crrrrrrrc}
    \toprule
        Attribute   &  $m\left( {\left\{ a \right\}} \right)$    & $m\left( {\left\{ b \right\}} \right)$       & $m\left( {\left\{ c \right\}} \right)$      & $m\left( {\left\{ a,b \right\}} \right)$      & $m\left( {\left\{ a,c \right\}} \right)$      & $m\left( {\left\{ b,c \right\}} \right)$       & $m\left( {\left\{ a,b,c \right\}} \right)$      \\
    \midrule
    SL    & $0.680{\rm{ + }}\frac{{{\rm{0}}{\rm{.126}}}}{{\rm{7}}}$       & $\frac{{{\rm{0}}{\rm{.126}}}}{{\rm{7}}}$     & $\frac{{{\rm{0}}{\rm{.126}}}}{{\rm{7}}}$      & $0.061{\rm{ + }}\frac{{{\rm{0}}{\rm{.126}}}}{{\rm{7}}}$      & $\frac{{{\rm{0}}{\rm{.126}}}}{{\rm{7}}}$      & $\frac{{{\rm{0}}{\rm{.126}}}}{{\rm{7}}}$      & $0.133{\rm{ + }}\frac{{{\rm{0}}{\rm{.126}}}}{{\rm{7}}}$      \\
    SW    & $0.503{\rm{ + }}\frac{{{\rm{0}}{\rm{.487}}}}{{\rm{7}}}$      & $\frac{{{\rm{0}}{\rm{.487}}}}{{\rm{7}}}$      & $\frac{{{\rm{0}}{\rm{.487}}}}{{\rm{7}}}$      & $\frac{{{\rm{0}}{\rm{.487}}}}{{\rm{7}}}$      & $0.{\rm{010 + }}\frac{{{\rm{0}}{\rm{.487}}}}{{\rm{7}}}$      & $\frac{{{\rm{0}}{\rm{.487}}}}{{\rm{7}}}$      & $\frac{{{\rm{0}}{\rm{.487}}}}{{\rm{7}}}$      \\
    PL    & $0.{\rm{920 + }}\frac{{{\rm{0}}{\rm{.080}}}}{{\rm{7}}}$      & $\frac{{{\rm{0}}{\rm{.080}}}}{{\rm{7}}}$     & $\frac{{{\rm{0}}{\rm{.080}}}}{{\rm{7}}}$      & $\frac{{{\rm{0}}{\rm{.080}}}}{{\rm{7}}}$      & $\frac{{{\rm{0}}{\rm{.080}}}}{{\rm{7}}}$      & $\frac{{{\rm{0}}{\rm{.080}}}}{{\rm{7}}}$      & $\frac{{{\rm{0}}{\rm{.080}}}}{{\rm{7}}}$      \\
    PW    & $0.{\rm{865 + }}\frac{{{\rm{0}}{\rm{.135}}}}{{\rm{7}}}$      & $\frac{{{\rm{0}}{\rm{.135}}}}{{\rm{7}}}$      & $\frac{{{\rm{0}}{\rm{.135}}}}{{\rm{7}}}$      & $\frac{{{\rm{0}}{\rm{.135}}}}{{\rm{7}}}$      & $\frac{{{\rm{0}}{\rm{.135}}}}{{\rm{7}}}$      & $\frac{{{\rm{0}}{\rm{.135}}}}{{\rm{7}}}$      & $\frac{{{\rm{0}}{\rm{.135}}}}{{\rm{7}}}$      \\
    \bottomrule
    \end{tabular}}%
  \label{tab:BPAs}%
\end{table}%

The Dempster combination rule is used to fuse the evidences in Table \ref{tab:BPAs}. Then, the final fusion results are shown in Table \ref{tab:result}.

\begin{table}[htbp]
  \centering
  \caption{The fusion result of the test sample in the closed world}
  \resizebox{\textwidth}{7mm}{
    \begin{tabular}{cccccccc}
    \toprule
   BPA &  $m\left( {\left\{ a \right\}} \right)$     & $m\left( {\left\{ b \right\}} \right)$       & $m\left( {\left\{ c \right\}} \right)$       & $m\left( {\left\{ a,b \right\}} \right)$       & $m\left( {\left\{ a,c \right\}} \right)$       &  $m\left( {\left\{ b,c \right\}} \right)$      & $m\left( {\left\{ a,b,c \right\}} \right)$  \\
    \midrule
   Value & 0.9995 & 0.0002 & 0.0001 & 0.0001 & 0     & 0     & 0 \\
    \bottomrule
    \end{tabular}}%
  \label{tab:result}%
\end{table}%

From the fusion results in Table \ref{tab:result}, it can be seen that as there is no interference from external environmental factors, this method assigns higher belief value than that in the open world ($m\left( {\left\{ a \right\}} \right) = 1$) to the correct class a, reflecting the effectiveness of this method in the complete frame of discernment.
In this section, It is proved that this method has good downward compatibility.
In the closed world where is no Interference caused by external environmental factors, the results obtained by this method are more discerning.
For practical applications in unstable environments (referring to switching between the open and closed world), just the last step of this method is needed to switch easily, which shows the simplicity and flexibility of this method.

\subsection{Robust experiments with adjusted the training set proportions in the closed world}
\label{sec:5.2}
In this section, the impact on classification performance is explored when changing the proportion of the training set in the closed world.
If the result of this method still has a good effect in the case of a low proportion of training, it indicates that this method has strong robustness.
We utilize the data set of $Iris$, and use 10 times randomized leave-out method respectively under the different proportions of the training set, to obtain the average value as the final recognition accuracy.

\begin{figure}
  \centering
  \includegraphics[scale=0.05]{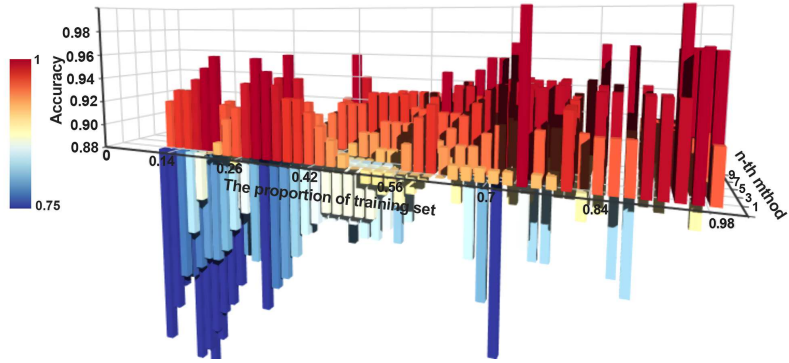}\\
  \caption{The recognition accuracy of different proportions of the training set in the closed world}\label{fig:accuracy}
\end{figure}

Table \ref{tab:allacc} shows all the accuracies obtained by this method at different proportions of training set, where  each column represents the proportion of the training set, and each row represents the n-th leave-out method.
Figure \ref{fig:accuracy} visualizes the data in the Table \ref{tab:allacc}, where the abscissa represents the proportion of the training set, the ordinate represents the $n$-th leave-out method, and the height of the cylinder indicates the recognition accuracy. It is clear from Figure \ref{fig:accuracy} that with the increase in the proportion of the training set, the recognition accuracy is getting higher and higher generally. Table \ref{tab:average} shows the average recognition accuracy under different training set proportions.

\begin{table}[htbp]
  \centering
  \caption{The recognition accuracy for the 1st, 2nd, ..., and 10th randomized leave-out experiment with different proportions of training set in the closed world}
  \resizebox{\textwidth}{32mm}{
    \begin{tabular}{ccccccccccc}
    \toprule
    Training part & 1st   & 2nd   & 3rd   & 4th   & 5th   & 6th   & 7th   & 8th   & 9th   & 10th \\
    \midrule
    10\%  & 70.83\% & 91.67\% & 75.00\% & 79.17\% & 66.67\% & 91.67\% & 62.50\% & 75.00\% & 87.50\% & 87.50\% \\
    20\%  & 87.50\% & 95.83\% & 79.17\% & 91.67\% & 75.00\% & 87.50\% & 87.50\% & 87.50\% & 87.50\% & 91.67\% \\
    30\%  & 95.83\% & 75.00\% & 87.50\% & 83.33\% & 87.50\% & 95.83\% & 87.50\% & 87.50\% & 87.50\% & 79.17\% \\
    40\%  & 91.67\% & 83.33\% & 83.33\% & 83.33\% & 87.50\% & 87.50\% & 91.67\% & 91.67\% & 79.17\% & 95.83\% \\
    50\%  & 86.67\% & 84.44\% & 82.22\% & 88.89\% & 91.11\% & 86.67\% & 93.33\% & 80.00\% & 93.33\% & 86.67\% \\
    60\%  & 88.33\% & 93.33\% & 88.33\% & 90.00\% & 91.67\% & 95.00\% & 90.00\% & 91.67\% & 93.33\% & 90.00\% \\
    70\%  & 88.89\% & 75.00\% & 91.67\% & 94.44\% & 91.67\% & 91.67\% & 88.89\% & 94.44\% & 0.9722 & 91.67\% \\
    72\%  & 88.89\% & 88.83\% & 86.11\% & 86.11\% & 91.67\% & 94.44\% & 97.22\% & 88.89\% & 91.67\% & 94.44\% \\
    74\%  & 100.00\% & 83.33\% & 94.44\% & 86.11\% & 94.44\% & 91.67\% & 80.56\% & 94.44\% & 88.89\% & 88.89\% \\
    76\%  & 88.89\% & 91.67\% & 94.44\% & 91.67\% & 88.89\% & 88.89\% & 80.56\% & 94.44\% & 91.67\% & 86.11\% \\
    78\%  & 88.89\% & 88.89\% & 91.67\% & 88.89\% & 94.44\% & 86.11\% & 94.44\% & 91.67\% & 94.44\% & 94.44\% \\
    80\%  & 93.33\% & 93.33\% & 93.33\% & 86.67\% & 86.67\% & 93.33\% & 86.67\% & 93.33\% & 0.8667 & 93.33\% \\
    82\%  & 86.11\% & 91.67\% & 88.89\% & 88.89\% & 86.11\% & 88.89\% & 88.89\% & 88.89\% & 94.44\% & 94.44\% \\
    84\%  & 91.67\% & 94.44\% & 86.11\% & 91.67\% & 94.44\% & 94.44\% & 88.89\% & 86.11\% & 97.22\% & 91.11\% \\
    86\%  & 91.67\% & 94.44\% & 91.67\% & 88.89\% & 88.89\% & 80.56\% & 94.44\% & 91.67\% & 86.11\% & 91.67\% \\
    88\%  & 91.67\% & 88.89\% & 80.56\% & 91.67\% & 97.22\% & 80.56\% & 97.22\% & 94.44\% & 0.8333 & 86.11\% \\
    90\%  & 94.44\% & 94.44\% & 91.67\% & 88.89\% & 91.67\% & 86.11\% & 94.44\% & 86.11\% & 88.89\% & 91.67\% \\
    92\%  & 94.44\% & 94.44\% & 88.89\% & 91.67\% & 86.11\% & 88.89\% & 88.89\% & 91.67\% & 88.89\% & 91.67\% \\
    94\%  & 94.44\% & 100.00\% & 97.22\% & 91.67\% & 91.67\% & 91.67\% & 91.67\% & 91.67\% & 91.67\% & 91.67\% \\
    96\%  & 97.22\% & 86.11\% & 91.67\% & 88.89\% & 97.22\% & 91.67\% & 91.67\% & 94.44\% & 94.44\% & 91.67\% \\
    98\%  & 91.67\% & 97.22\% & 88.89\% & 88.89\% & 91.67\% & 88.89\% & 94.44\% & 97.22\% & 88.89\% & 88.89\% \\
    \bottomrule
    \end{tabular}}%
  \label{tab:allacc}%
\end{table}%

\begin{table}[htbp]
  \centering
  \caption{The average accuracy of leave-out method in the closed world}
  \resizebox{\textwidth}{8mm}{
    \begin{tabular}{cccccccccccc}
    \toprule
    Training part & 10\%  & 20\%  & 30\%  & 40\%  & 50\%  & 60\%  & 70\%  & 72\%  & 74\%  & 76\%  & 78\% \\
    \midrule
    Average accuracy & 78.75\% & 87.08\% & 86.67\% & 87.50\% & 87.33\% & 91.17\% & 90.57\% & 90.28\% & 90.28\% & 89.72\% & 91.39\% \\
    \midrule
    Training part & 80\%  & 82\%  & 84\%  & 86\%  & 88\%  & 90\%  & 92\%  & 94\%  & 96\%  & 98\%  &  \\
    \midrule
    Average accuracy & 90.67\% & 89.72\% & 91.11\% & 90.00\%   & 89.17\% & 90.83\% & 90.56\% & 93.33\% & 92.50\% & 91.67\% &  \\
    \bottomrule
    \end{tabular}}%
  \label{tab:average}%
\end{table}%

As known, the accuracy of classification methods is positively related to the proportion of the training set.
When the training set contains most samples, the model will cause over-fitting, leading to the evaluation result that may not be stable and accurate, which is also the disadvantage of using the leave-out method.
In order to cross-validate the accuracy of this method, the cross-validation method is used to modify the average accuracy in Table \ref{tab:average}.
Compared with the leave-out method, the cross-validation method can improve the stability and fidelity of the evaluation results to a certain extent.
Taking into account that the proportion that is not a multiple of ten cannot divide the data set $D$ into $k$ subsets evenly, which makes the process of evaluation difficult, so 10-times-10-fold cross-validation method and 10-times-5-fold cross-validation method are used respectively for cases with 80\% and 90\%  of the proportion of the training sets. The calculated accuracy also represents the accuracy of its surrounding interval [-5,5]. For example, 10-times-5-fold cross-validation method is as follows:
\paragraph{1)}Divide the $Iris$ data set $D$ into 5 mutually exclusive subsets of the same size. In other words, let $D = {D_1} \cup {D_2} \cup {D_3} \cup {D_4} \cup {D_5}$. ${D_1},{D_2},{D_3},{D_4},{D_5}$ are mutually exclusive, where each subset is obtained from D by a stratified sampling method, thereby maintaining the consistency of data.

\paragraph{2)}Use 4 subsets as the training set and the remaining one as the test set each time. In this way, $k$ sets of training / test sets can be obtained, so that $k$ training and testing can be performed, and the average value of $k$ test results can be finally obtained.
\paragraph{3)}Repeat step 2) 10 times to get the average value. The purpose of this step is to reduce the differences caused by different sample partitions.
\paragraph{4)}Modify the original data from 76\% to 94\% according to the accuracy correction formula, which is defined as follows:
\begin{equation}
accuracy = \frac{{accurac{y_{leave - out}} + accurac{y_{cross - validation}}}}{2}
\end{equation}

The results of using 10 times cross-validation for 80\% and 90\% of the training set proportions are shown in Table \ref{tab:cross-validate}, where each row represents the proportion of the training set, and each column represents the $n$-th cross-validation.
\begin{table}[htbp]
  \centering
  \caption{The accuracy of cross-validation in the closed world for the 1st, 2nd, ..., and 10th randomized leave-out experiment}
  \resizebox{\textwidth}{5mm}{
    \begin{tabular}{cccccccccccc}
    \toprule
    Training part & 1st & 2nd & 3rd & 4th & 5th & 6th & 7th & 8th & 9th & 10th & Average accuracy \\
    \midrule
    80\%  & 92.00\%  & 92.00\%  & 89.33\% & 91.33\% & 89.33\% & 91.33\% & 90.67\% & 90.00\%   & 91.33\% & 90.67\% & 90.80\% \\
    90\%  & 91.33\% & 92.00\%  & 92.00\%  & 91.33\% & 91.33\% & 90.67\% & 92.00\%  & 91.33\% & 91.33\% & 92.00\%  & 91.53\% \\
    \bottomrule
    \end{tabular}}%
  \label{tab:cross-validate}%
\end{table}%

After using the accuracy correction formula, the corrected accuracy of  the proportion of the training set is as follows:
\begin{table}[htbp]
  \centering
  \caption{The modified accuracy with different proportion of the training set }
  \resizebox{\textwidth}{5mm}{
    \begin{tabular}{ccccccccccc}
    \toprule
    Training part & 76\%  & 78\%  & 80\%  & 82\%  & 84\%  & 86\%  & 88\%  & 90\%  & 92\%  & 94\% \\
    \midrule
    Modified accuracy & 90.26\% & 91.09\% & 90.73\% & 90.26\% & 90.95\% & 90.76\% & 90.35\% & 91.18\% & 91.05\% & 92.43\% \\
    \bottomrule
    \end{tabular}}%
  \label{tab:modified}%
\end{table}%

\begin{figure}
  \centering
  \includegraphics[scale=0.35]{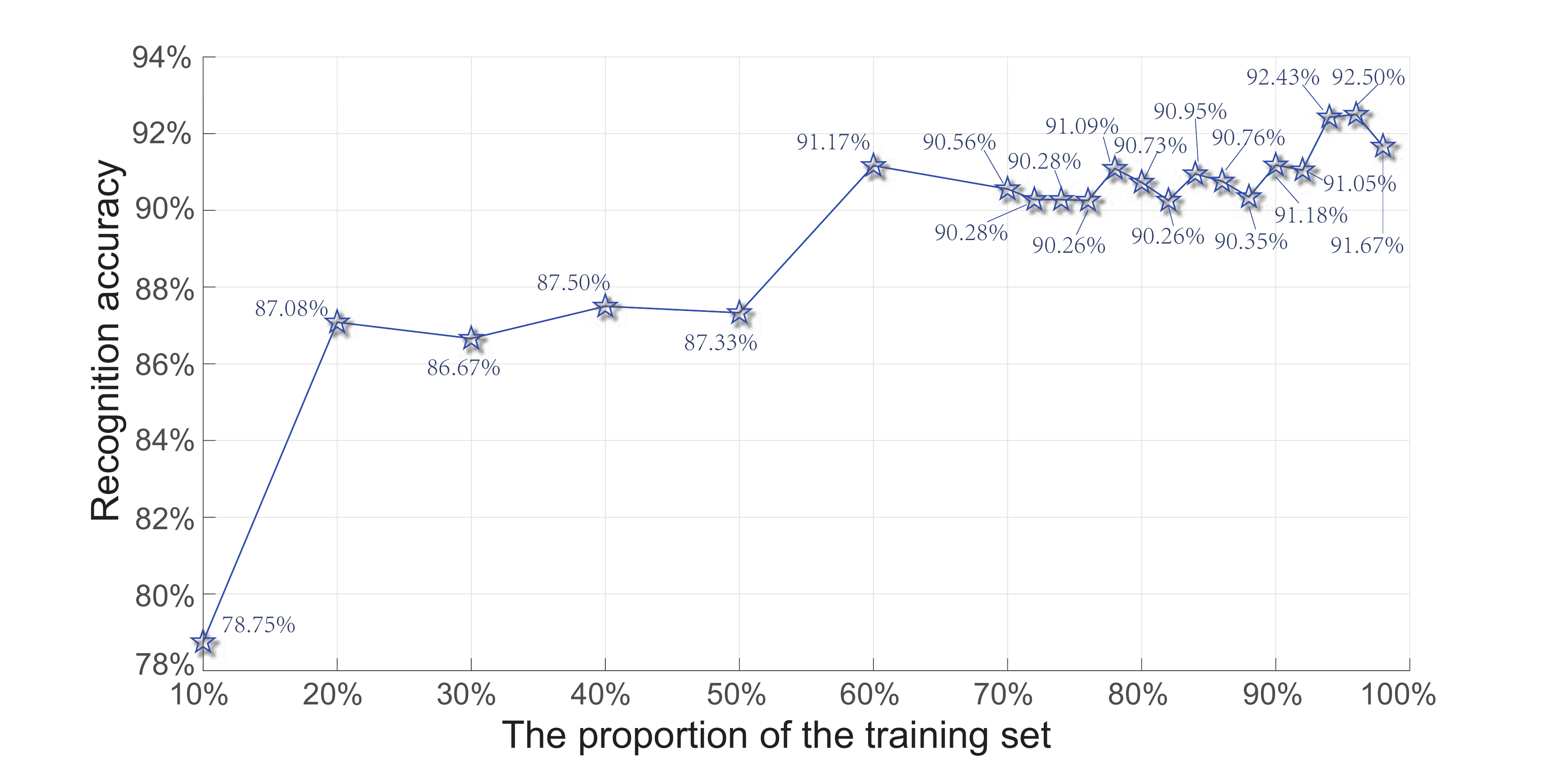}\\
  \caption{The recognition accuracy}\label{fig:recognition}
\end{figure}

Figure \ref{fig:recognition} shows the classification accuracy of this method under different proportions of the training set. The abscissa represents the proportion of training set and the ordinate represents the recognition accuracy.
It can be roughly seen from Figure \ref{fig:recognition} that when the proportion of the training set is higher, the recognition accuracy is higher, and finally tends to stabilize gradually. It is showed that when the number of training samples is sufficient, this method can correctly recognize unknown classes with high accuracy.
Even when the number of training samples is not enough, as shown in Figure \ref{fig:recognition} where the training set accounts for only 10\%, the accuracy of recognition accuracy approaches 80\%, and when it accounts for 20\%, the recognition accuracy can reach 87\%. This method is of great help in practical engineering applications with little experimental data.
This section verifies that this method has high efficiency and strong robustness.

\subsection{Comparative analysis of experimental results of improved Dempster combination rule}
\label{sec:5.3}
Although the recognition accuracy of this method is relatively satisfactory in the closed world,
there is still flaw when using the Dempster combination rule to fuse evidence.
The reason for the flaw is that the Dempster combination rule is sensitive to $m\left( {\left\{ X \right\}} \right) = 0\left( {X \in {2^\Theta }} \right)$ during fusion.
Here is a special example to explain the impact of the flaw.
Assume there are 2 sensors to recognize 3 classes A, B, and C. The data obtained from the sensors are as follows:
\begin{equation}
\begin{array}{l}
{m_1}\left( {\left\{ A \right\}} \right) = 0.8,{m_1}\left( {\left\{ B \right\}} \right) = 0,{m_1}\left( {\left\{ C \right\}} \right) = 0.2,\\
{m_2}\left( {\left\{ A \right\}} \right) = 0,{m_2}\left( {\left\{ B \right\}} \right) = 0.8,{m_2}\left( {\left\{ C \right\}} \right) = 0.2,
\end{array}
\end{equation}
The results obtained from Dempster combination rule are as follows:
\begin{equation}
m\left( {\left\{ A \right\}} \right) = 0,m\left( {\left\{ B \right\}} \right) = 0,m\left( {\left\{ C \right\}} \right) = 1,
\end{equation}

Obviously, using Dempster's combination rule will produce result that is counter-intuitive, which is very unreasonable and unreliable.
It can be seen that the classic Dempster combination rule cannot work when the evidence has a highly conflict. However, the negation method can provide information from the opposite perspective. When using negation of BPA instead of the original BPA, the following data can be obtained:
\begin{equation}
\begin{array}{l}
{{\bar m}_1}\left( {\left\{ A \right\}} \right) = 0.1,{{\bar m}_1}\left( {\left\{ B \right\}} \right) = 0.5,{{\bar m}_1}\left( {\left\{ C \right\}} \right) = 0.4,\\
{{\bar m}_2}\left( {\left\{ A \right\}} \right) = 0.5,{{\bar m}_2}\left( {\left\{ B \right\}} \right) = 0.1,{{\bar m}_2}\left( {\left\{ C \right\}} \right) = 0.4,
\end{array}
\end{equation}
Using the Dempster combination rule on the above data can get the following results:
\begin{equation}
\bar m\left( {\left\{ A \right\}} \right) = 0.19,\bar m\left( {\left\{ B \right\}} \right) = 0.19,\bar m\left( {\left\{ C \right\}} \right) = 0.62
\end{equation}

From the above fusion results, it can be seen that the belief value of non-target C is higher than that of non-target A and non-target B, and the belief value of non-target A is equal to that of non-target B, which is in common sense.
Therefore, in the closed world, the recognition accuracy in this paper can be further improved through the negation of BPA.

\section{Conclusions}
\label{sec:6}
In the framework of Dempster-Shafer evidence theory, this paper proposes a method that can generate both GBPA in the open world and BPA in the closed world.
The proposed method is based on a multi-subset propositional representation model, where the principle is  simple and intuitive as well as be able to cope with complex and unstable environmental conditions.
The proposed method enhances the ability to model incomplete information in the open world, thus, it can detect more unknown information in the external environment and to reduce the loss of effective information.
Based on the experiments of the data set of $Iris$ in the open world, the experimental result shows that the fusion result of the proposed method can be very distinguishable, and its recognition accuracy in practical applications also has a promising performance.
We discussed the  property of the proposed method, and verified that the proposed method has a  downward compatibility  in classification and robust experiments.
In the closed world, the proposed method can assign a belief value of above 0.999 to the correct class according to the fusion result.
Even if the system is in an unfamiliar environment with a small number of training samples, the robustness experiment that investigates the recognition accuracy by changing the proportion of the training set verifies that the recognition accuracy can be as high as 87\% even when the proportion of the training set is 20\%.

%

\bibliographystyle{unsrt} 

\bibliography{References}


\end{document}